\documentclass{article}

\include{definitions.tex}

\usepackage[final]{nips_2018}

\usepackage[utf8]{inputenc} 
\usepackage[T1]{fontenc}    
\usepackage{hyperref}       
\usepackage{url}            
\usepackage{booktabs}       
\usepackage{amsfonts}       
\usepackage{nicefrac}       
\usepackage{microtype}      

\usepackage{subfig}
\usepackage[inline]{enumitem}
\usepackage{floatrow}
\usepackage{amsmath}
\usepackage{amssymb}
\usepackage{graphicx}
\usepackage{array}
\usepackage{algorithm}
\usepackage{makecell}
\usepackage[noend]{algpseudocode}
\makeatletter
\def\BState{\State\hskip-\ALG@thistlm}
\makeatother
\newcolumntype{M}[1]{>{\centering\arraybackslash}m{#1}}
\usepackage[export]{adjustbox}
\usepackage{caption}
\usepackage{fancyhdr}
\newcommand{\sfigmediummedium}[0]{7cm}
\newcommand{\sfigmediumsmall}[0]{5.8cm}

\title{Fast On-the-fly Retraining-free Sparsification \\ of Convolutional Neural Networks}

\author{\normalfont{Amir H.\ Ashouri}\\ 
University of Toronto\\
Canada\\
{\tt\small aashouri@ece.utoronto.ca}
\and
Tarek S.\ Abdelrahman\\
University of Toronto\\
Canada\\
{\tt\small tsa@ece.utoronto.ca}
\and
Alwyn Dos Remedios\\
Qualcomm Inc. \\
Canada\\
{\tt\small adosreme@qti.qualcomm.com}
}

\begin{document}

\maketitle
\thispagestyle{fancy}

\begin{abstract}
 Modern Convolutional Neural Networks (CNNs) are complex, encompassing millions of parameters. Their deployment exerts computational, storage and energy demands, particularly on embedded platforms. Existing approaches to prune or {\em sparsify} CNNs require retraining to maintain inference accuracy. Such retraining is not feasible in some contexts. In this paper, we explore the sparsification of CNNs by proposing three model-independent methods. Our methods are applied on-the-fly and require no retraining. 
 We show that the state-of-the-art models' weights can be reduced by up to 73\% (compression factor of 3.7$\times$) without incurring more than 5\% loss in Top-5 accuracy. Additional fine-tuning gains only 8\% in sparsity, which indicates that our fast on-the-fly methods are effective.
\end{abstract}

\section{Introduction}

There has been a significant growth in the number of parameters (i.e., layer weights), and the corresponding number of multiply-accumulate operations (MACs), in state-of-the-art CNNs \cite{LeCun1998,AlexNet-Krizhevsky2012,vgg-16,inceptionV1-Szegedy2015,resnet,squeezNet,inceptionv3,inceptionV4Szegedy2016}. For example, the number of parameters/MACs has increased from 2.6K/341K in LeNet-5 to 23.5M/3.9G in ResNet-50~\cite{Sze2017}. Such growth makes it difficult to deploy CNNs, particularly on embedded devices, where memory capacity, computational power, and energy are constrained. 

Thus, it is to no surprise that several techniques exist for ``pruning'' or  ``sparsifying'' CNNs (i.e., forcing some model weights to 0) to both compress the model and to save computations during inference. 
Examples of these techniques include: iterative pruning and retraining~(\cite{Cun1990,Hassibi93,Dong2017,Sun2016,Mathew}), Huffman coding~(\cite{Han}), exploiting granularity~(\cite{Mao2017,Han2016}), structural pruning of network connections~(\cite{Wen2016,Mao,Anwar2015,Niculae2017}), and Knowledge Distillation (KD) (\cite{hinton2015distilling}). 

A common theme to the aforementioned techniques is that they require a retraining of the model to fine-tune the remaining non-zero weights and maintain inference accuracy.  Such retraining, while feasible in some contexts, is not feasible in others, particularly industrial ones. For example, for mobile platforms, a machine learning model is typically embedded within an app for the platform that the user directly downloads. The app utilizes the vendor's platform runtime support (often in the form of a library) to load and use the model. Thus, the platform vendor must sparsify the model at runtime, i.e., {\em on-the-fly}, within the library with no opportunity to retrain the model. Further, the vendor rarely has access to the labelled data used to train the model. While techniques such as Knowledge Distillation (\cite{hinton2015distilling}) can address this lack of access, it is not possible to apply it on-the-fly.

In this paper, we seek to answer some fundamental questions that relate to the trade-off between sparsity and inference accuracy: (1) To what extent a CNN can be sparsified without retraining while maintaining a reasonable inference accuracy, (2) What are good model-independent methods for sparsifying CNNs, and (3) Can such sparsification benefit from autotuning \cite{han2014automatic,Ashourisurvey}. We focus on sparsification leaving the actual exploitation of the resulting sparsity to future work.

We develop fast retraining-free  sparsification methods that can be deployed for on-the-fly sparsification of CNNs in the contexts described above. There is an inherent trade-off between sparsity and inference accuracy. Our goal is to develop {\em model-independent} methods that result in large sparsity with little loss to inference accuracy. 
We develop three model-independent sparsification methods: {\em flat}, {\em triangular}, and {\em relative}.
We implement these methods in TensorFlow and use the framework to evaluate the sparsification of several pretrained models: Inception-v3, MobileNet-v1, ResNet, VGG, and AlexNet. Our evaluation shows that up to 73\% of layer weights in some models may be forced to 0,  incurring only a 5\% loss in inference accuracy. While the relative method appears to be more effective for some models, the triangular method is more effective for others. Thus, predictive modeling is needed to identify, at run-time, the optimal choice of method and it hyper-parameters.

Thus, this paper makes the following contributions: 
 \begin{itemize}
 \item We propose and evaluate three model-independent sparsification methods for CNNs.
 \item We develop a framework sparsifying CNNs built around TensorFlow, which does not require retraining after sparsification and it does so on-the-fly. 
\end{itemize}

The remainder of the paper is organized as follows. We discuss related work in Section~\ref{related}, In Section~\ref{sparsification} we present our three sparsification methods and the overall framework. 
Section~\ref{EXresults} presents our experimental evaluation. Finally, in Section~\ref{sec:conc} we offer some concluding remarks.

\section{Related Work}
\label{related}

Generally, this research can be classified into three broad categories: (1)~pruning and weight sparsifying ~\cite{Mao2017,Mathew,Sun2016,Cun1990,Parashar2017,Han}, (2)~structural pruning ~\cite{Wen2016,Mao,Anwar2015,Niculae2017}, and (3)~low rank approximation ~\cite{Zhang,Musco2017,Liu,Srivastava2014}. Nearly all of this work uses retraining to fine-tune the resulting sparsified, pruned or reduced model~\cite{Srinivas,Yu2017}.

Le Cun et al.~\cite{Cun1990} proposed Optimal Brain Damage to reduce neural connection of pairs using the \emph{saliency} of model parameters. Others~\cite{Hassibi93,Dong2017} extend this work to use second order derivatives. More recently, Mao et al.~\cite{Mao2017,Mao,Hana} explore coarse-grain and fine-grain pruning and evaluate the trade-off between accuracy and sparsity using recent CNNs.

Huang et al. \cite{Huang2018} proposed a try-and-learn algorithm for pruning redundant filters in CNNs. They use a reward function to aggressively prune with minimal loss of accuracy; however, their method requires retraining as well as user input. Earlier, 

Sun et al.~\cite{Sun2016} proposed ConvNets as a framework that can be used to iteratively learn sparsified neural connections through correlations among neural activations. The connections are dropped iteratively one layer at a time and the model is retrained.

Mathew et al.~\cite{Mathew} proposed a framework that compensates for the loss of accuracy after sparsification by retraining. They  quantize their sparsified model for an embedded architecture and observe a nearly 4$\times$ improvement in the inference speed with 80\% sparsity.

Wen et al.~\cite{Wen2016} propose SSL: a sparsifying framework to exploit and regularize structural sparsity of a sample DNN. Their evaluation using ResNet reduces a few layers while improving inference accuracy by around 1.5\%. 

Parashar et al.~\cite{Parashar2017} propose the SCNN accelerator architecture that utilizes a  compressed encoding of sparse weights and activations. They observe up to 2.7$\times$ energy improvement during both re-training and inference.

In contrast to the above work, our method limits itself to the sparsity introduced into model weights (as opposed to activations) and avoid retraining. Our approach results in sparse models without the overhead of retraining and without the need for training data. This is particularly important in some industrial settings, access to training data is not possible and the retraining represents a significant time investment, thus, the sparsification needs to be on-the-fly.

\section{Sparsification Methods}
\label{sparsification}

Sparsity in a CNN stems from three main sources: (1)  weights within convolution (\texttt{Conv}) and fully-connected (\texttt{FC}) layers (some of these weights may be zero or may be forced to zero); (2) activations of layers, where the often-applied \texttt{ReLU} operation results in many zeros (\cite{Sze2017}); and (3) input data, which may be sparse. In this paper, we focus on the first source of sparsity, in both \texttt{Conv} and fully connected layers. This form of sparsity can be determined a priori, which alleviates the need for specialized hardware accelerators.

\begin{figure}[t!]
\centering
\begin{subfloat}[\texttt{Inception-v3} Softmax]{
    \label{fig:histoFC}
            \resizebox{\sfigmediumsmall}{!}{\includegraphics{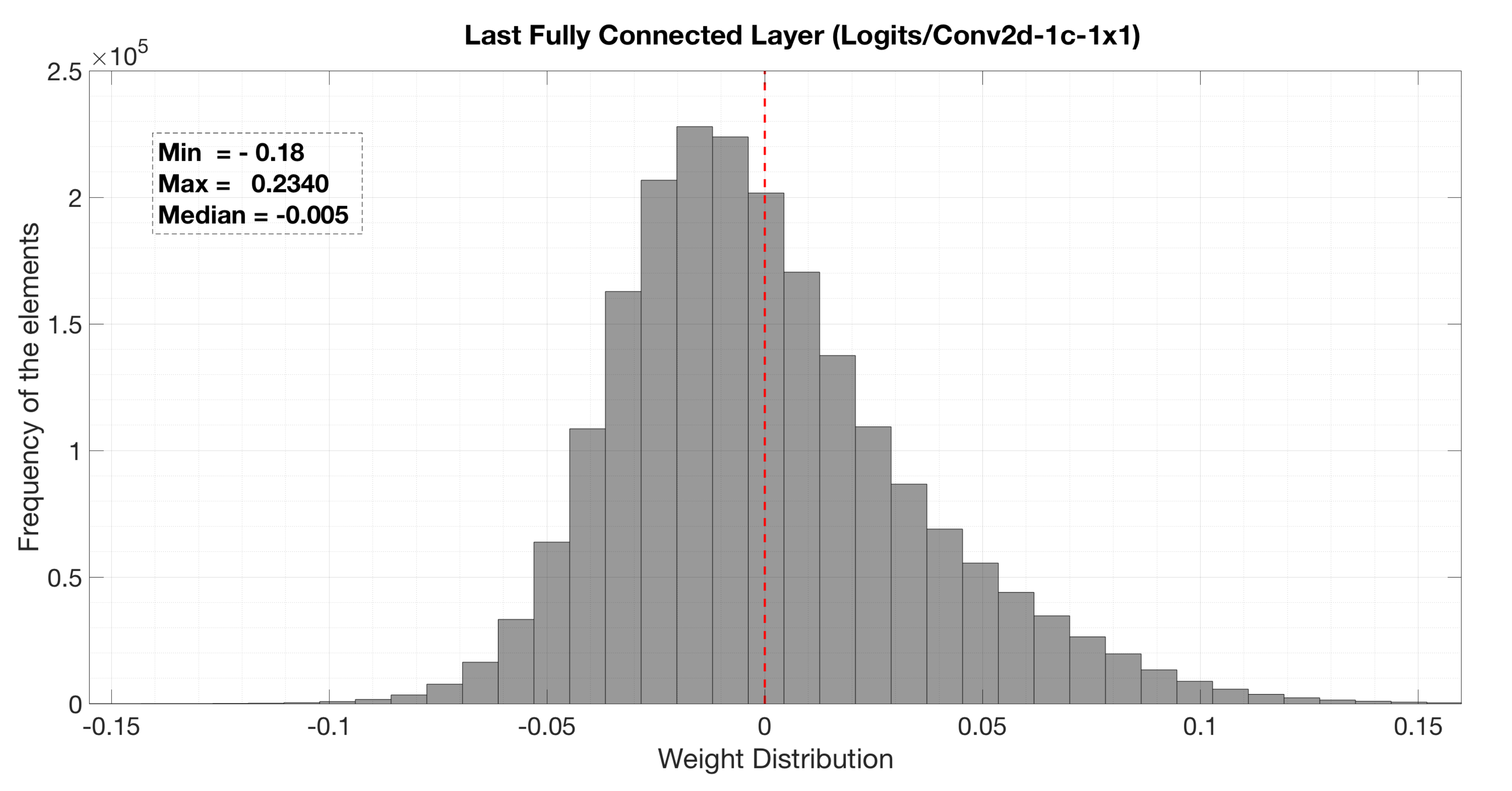}}
            }
     \end{subfloat}
\begin{subfloat}[\texttt{Inception-v3} Last Conv]{
    \label{fig:histoLLConv}
            \resizebox{\sfigmediumsmall}{!}{\includegraphics{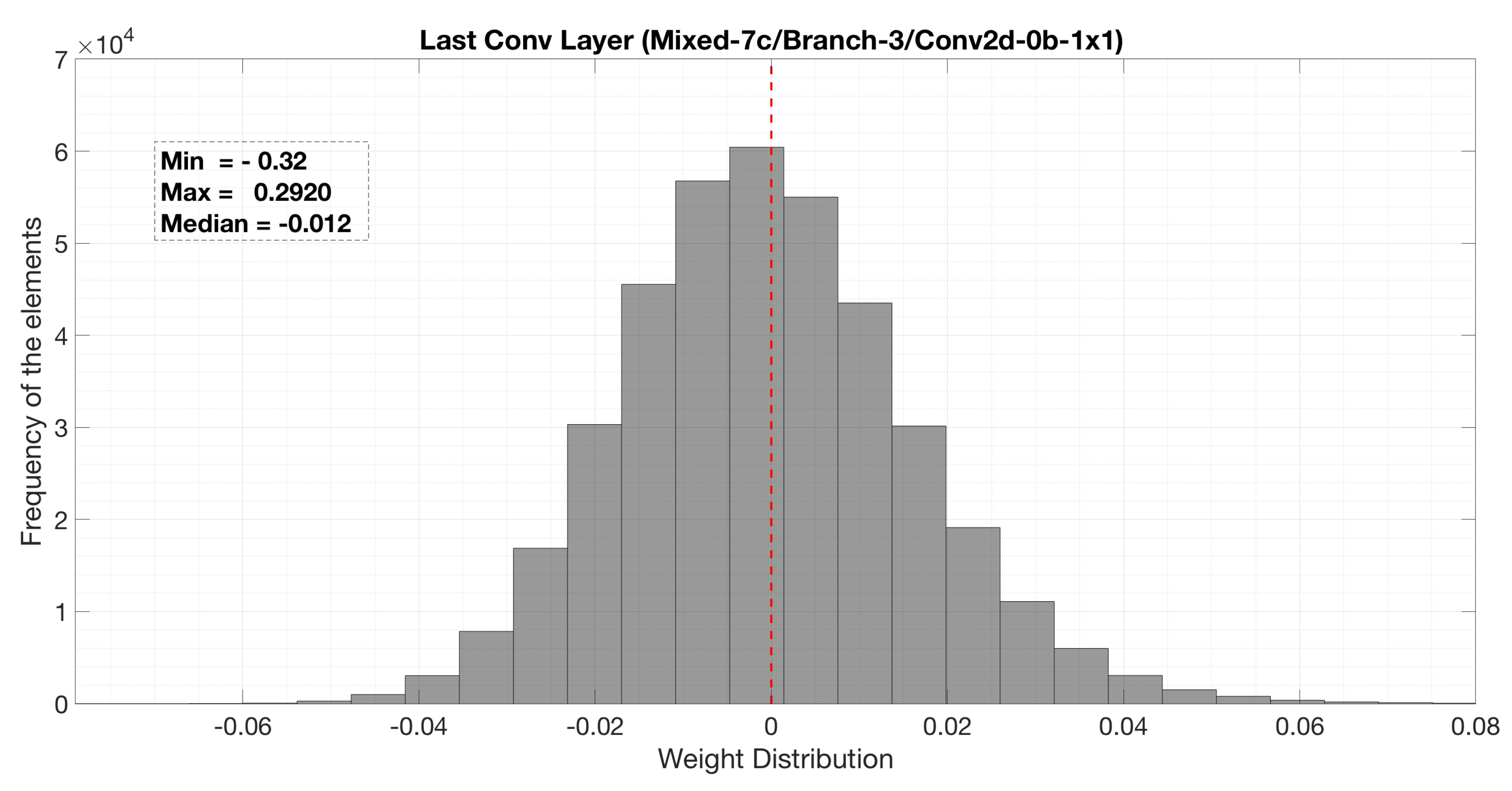}}
            }
     \end{subfloat}
\caption{Distribution of the weights in Inception-v3}
\label{fig:inceptionDistribution}
\end{figure}

The input to our framework is a CNN that has $L$ layers, numbered $1 \ldots L$.  The weights of each layer $l$
are denoted by $\omega_{l}$. In the case of convolution layers, these weights form a 4D tuple (tensor) $\omega_{l} \epsilon R^{{F_l}\times{C_l}\times{H_l}\times{W_l}}$, where ${F_l}, {C_l}, {H_l}$, and ${W_l}$ are the dimensions of the $l$-th  weight tensor across the different axes of filters, channels, height and width, respectively. 
We sparsify these weights  using a sparsification function, $\mathbb{S}$, which takes as input $\omega_{l}$ and a threshold $\tau_l$ from a vector of thresholds $T$. Each weight $i$ of $\omega_l$ is modified by $\mathbb{S}$ as follows:%

\begin{equation}
\small
\label{eq:sparsificationGeneral}
\mathbb{S}(\omega_{l}(i),\tau_{l}) = \begin{cases} 0 & \text{if} \ |{\omega_{l}(i)}| \leq \tau_{l} \\ \omega_{l}(i) & \text{otherwise} \ \end{cases}
\end{equation}

\noindent where $\tau_l = T(l)$ is the threshold used for layer $l$. Thus, applying a single threshold $\tau_l$ forces weights within $-\tau_l$ and $+\tau_l$ in value to become 0. 
We define the {\em sparsity ratio} for each layer, denoted by $s_l$ as the number of zero weights present after sparsification divided by the total number of weights in the layer. We similarly define the {\em model sparsity ratio}, denoted by $s_m$ as the total number of zero weights in all layers of the model divided by the total number of weights in the model.
Our use of thresholds to sparsify layers is motivated by the fact that recent CNNs' weights are distributed around the value $0$.
Figure~\ref{fig:inceptionDistribution} shows the histograms of the weights of the last fully-connected and convolution layers of the Inception-v3 pretrained model. The figure shows that the weights are distributed in a quasi-symmetric way around $0$. 
Thus, applying a single threshold $\tau_l$ forces weights within $-\tau_l$ and $+\tau_l$ in value to become 0.  

\begin{figure}[t!]
\centering
\subfloat[Sparsification Methods]{
    \label{fig:sparsificationMethods}
            \resizebox{\sfigmediummedium}{!}{\includegraphics{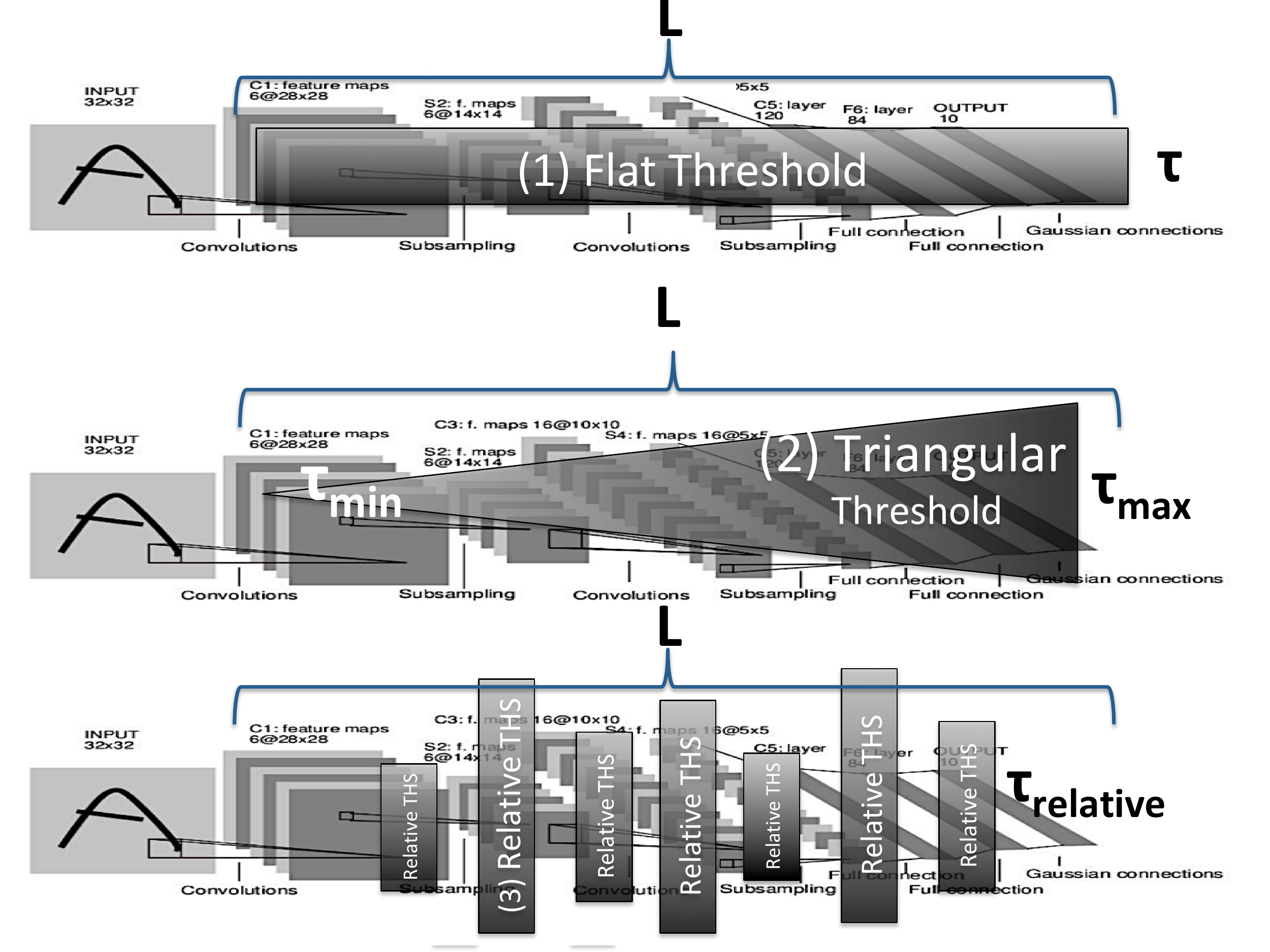}}
            }
\subfloat[Sparsification Workflow]{
    \label{fig:proposedMethodology}
            \resizebox{\sfigmediummedium}{!}{\includegraphics{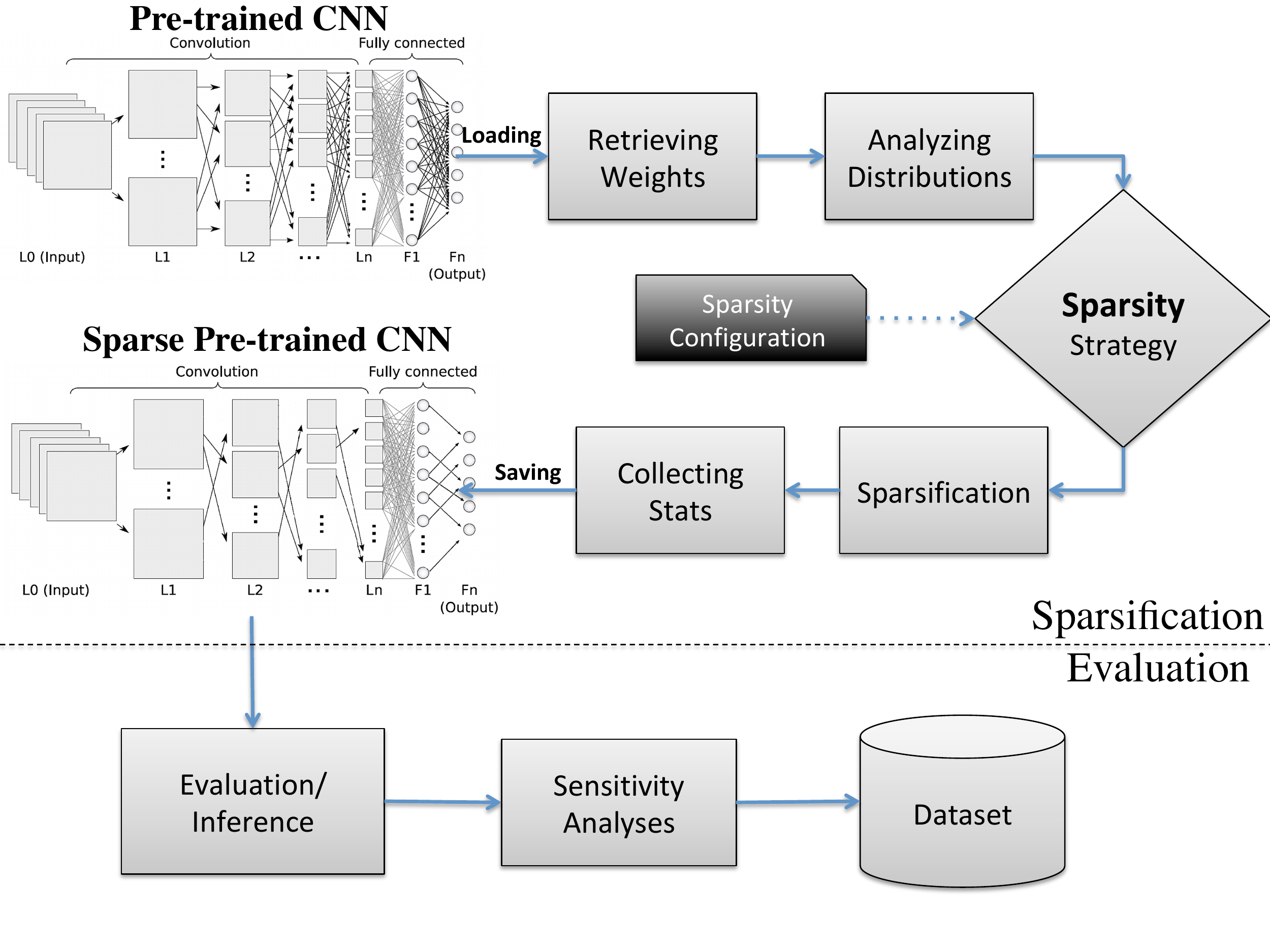}}
            }
\caption{Proposed Sparsification Framework}
\label{fig:proposedFramework}
\end{figure}

The choice of the values of the elements of the vector $T$ defines a {\em sparsification method}. These values impact the resulting sparsity and inference accuracy. We define and compare three sparsification methods. The {\em flat} method defines a constant threshold $\tau$ for all layers, irrespective of the distribution of their corresponding weights. The {\em triangular} method is inspired by the size variation of layers in some state-of-the-art CNNs, where the early layers have smaller number of parameters than latter layers. Finally, the {\em relative} method defines a unique threshold for each layer that sparsifies a certain percentage of the weights in the layer. The three methods are depicted graphically in Figure~\ref{fig:sparsificationMethods}. The high level work-flow of the sparsification framework is depicted in Figure \ref{fig:proposedMethodology}.

\subsection{Flat Method}
\label{sparsification:flat}

This method defines a constant threshold $\tau$ for all layers, irrespective of the distribution of their corresponding weights. It is graphically depicted in the top of Figure~\ref{fig:sparsificationMethods}.  The weights of the layers are profiled to determine the span $\sigma_{min} =\underset{\forall l in L}{min}(max(\omega_{l})-min(\omega_{l}))$. This span corresponds to the layer $k$ having the smallest range of weights within the pretrained model. This span is used as an upper-bound value for our flat threshold $\tau$. Since using $\sigma_{min}$ as a threshold eliminates all the weights in layer $k$ and is likely to adversely affect the accuracy of the sparsified model, we use a fraction $\delta$,  $0 \leq \delta \leq 1$, of the span: 
\begin{equation}
\small
\label{eq:sparsificationGeneral:flat}
\tau_{l} = \sigma_{min} \times {\delta}
\end{equation}
\noindent where $\delta$ is a parameter of the method that can be varied to achieve different degrees of model sparsity.

\subsection{Triangular Method}
\label{sparsification:triangular}

The triangular method is inspired by the size variation of layers in state-of-the-art CNNs: early \texttt{Conv} layers have a smaller number of parameters than in later layers, i.e., final \texttt{Conv} and \texttt{FC} layers. 

Thus, the triangular sparsification method introduces the highest sparsity in the final layers and gradually diminishes the sparsification towards the early layers. This is done using an isosceles triangular shape for the values of the thresholds, as shown in Figure~\ref{fig:sparsificationMethods}. This not only caters to the size variation of layers, it also avoids aggressive sparsification in early layers, which can adversely impact subsequent layers because pruning too many neurons in earlier layers can effectively remove many paths in layers further down.

The triangular method is defined by two thresholds $\tau_{min}$ and $\tau_{max}$ for respectively the first convolution layer (i.e, layer 1) and the last fully connected layer (i.e., layer $L$). They represent the thresholds at the tip and the base of the triangle in middle part of Figure~\ref{fig:sparsificationMethods}. These thresholds are determined by the span of the weights in each of the two layers. Thus,

\begin{equation}
  \begin{aligned}
    \tau_{min} = \sigma_{conv} * \delta_{conv}  \\
    \tau_{max} = \sigma_{fc} * \delta_{fc}
    \end{aligned}
\end{equation}

\noindent where $\sigma_{conv}$ is the span of the weights in the first convolution layer, defined in a similar way as for the flat method, and it represents an upper bound on $\tau_{min}$. Thus, $\delta_{conv}$ is a fraction that ranges between $0$ and $1$. Similarly, $\sigma_{fc}$ is the span of the weights in the last fully connected layer and it represents an upper bound on $\tau_{max}$. Thus, $\delta_{fc}$ is a fraction that ranges between $0$ and $1$.

The thresholds of the remaining layers are dictated by the position of these layers in the network. Thus, the $\tau$ value for each layer $l$, $1 \leq l \leq L$, is given by:
\begin{equation}
\small
\label{eq:sparsificationGeneral:triangular}
  \tau_{l} =     
  \begin{cases}                
    \tau_{min}                                     & \text{if}~~~~~  \  l = 1 \\  
    \tau_{max}                                     & \text{if}~~~~~  \  l = L \\
    \dfrac{\tau_{max}-\tau_{min}}{L} \times (l-2)  & \text{Otherwise} 
  \end{cases}
\end{equation}

\subsection{Relative Method}
\label{sparsification:relative}

Several state-of-the-art CNNs, such as Inception-v3~\cite{inceptionv3} or AlexNet~\cite{AlexNet-Krizhevsky2012} no longer exhibit a linear expansion in the design of their layers' parameters. Rather, they exhibit a more complex structure in which bigger layers are replaced by multiple smaller layers that are grouped together. To this end, a flat or triangular sparsification method might not be intuitive and efficient because it can aggressively prune smaller and under-sparsify wider layers. 

The relative method attempts to address this issue. It defines a unique threshold for each layer that is solely based on the distribution of the weights of that layer. 
The value of $\tau_l$ is picked such that it is the $\delta_l$th {\em percentile} of the weights in layer $l$. That is, $\tau_l$ is a value such that approximately $\delta_l$ of the weights (in absolute value) are less than or equal to $\tau_l$. Thus, our sparsification function (Equation~\ref{eq:sparsificationGeneral}) results in approximately $\delta_l$ zeros for layer $l$.

In particular, it uses the $\delta^{th}$ percentile of distribution's weight in layer $l$ denoted by $\delta_{l}$. 
Thus each element of the vector $T(l) = \tau_l$ is defined as:

\begin{equation}
\small
\label{eq:sparsificationGeneral:relative}
\tau_{l} = (max(\omega_{l})-min(\omega_{l})) \times {\delta_l}
\end{equation}
\noindent where $0 \leq \delta_l \leq 1$ defines the desired percentile of the $\delta_{l}$ of zero weights in each layers.

\section{Experimental Evaluation and Comparison}
\label{EXresults}

Figure \ref{fig:proposedMethodology} depicts the highlevel flow of our proposed sparsification framework. A pretrained CNN is first loaded, as shown on the upper left side of the figure. The meta-graph and the weights of the model are then retrieved. Next, the histogram of the weights in each layer is generated, and the threshold value(s) of the sparsification method used are determined. Once the method is applied, the sparsified weights are saved as a new sparsified version of the CNN. The sparsified model is then evaluated to capture its new inference accuracy. 
Note that on an industrial use case such as an app on a Mobile device, the pretrained models are embedded into the app and only communicate with the API using a library. Thus, our method can also be applied on-the-fly and sparsifies the loaded weights as they are appended to the device memory and just before the execution.

We evaluate our sparsification methods using TensorFlow v1.4 with CUDA runtime and driver v8. The evaluation of Top-5 accuracy is done on an NVIDIA's GeForce GTX 1060 with a host running Ubuntu 14.04, kernel v3.19 using ImageNet (\cite{imagenet}).

\paragraph{Metrics}
\label{metrics}

We evaluate our sparsified models using two key metrics. The first is the model sparsity ratio, as defined in Section~\ref{sparsification}. We use it as a proxy for the amount of memory and computation (i.e, number of MACs) savings possible when sparsity is exploited. While the savings will not equal the sparsity ratio (because of compression overheads), the extent to which these savings introduce remains proportional to the ratio.
does not explicitly reflect the quantitative value saved by a sparsified model, nevertheless, it hints the compression's potential of a model to be exploited later by canceling the multiplications and addition of 0s. 
The second metric is the {\em drop in inference accuracy}. The acceptable drop in inference accuracy is highly application dependent. In our evaluation, we set the acceptability threshold for normalized inference accuracy to 95\%. We explore sparsification methods that result in the largest model sparsity within this threshold.

\begin{figure}[!t]
\centering
\subfloat[Inception-v3]{
\label{fig:proposedSparsificationsL30}
\resizebox{0.46\textwidth}{!}{\includegraphics{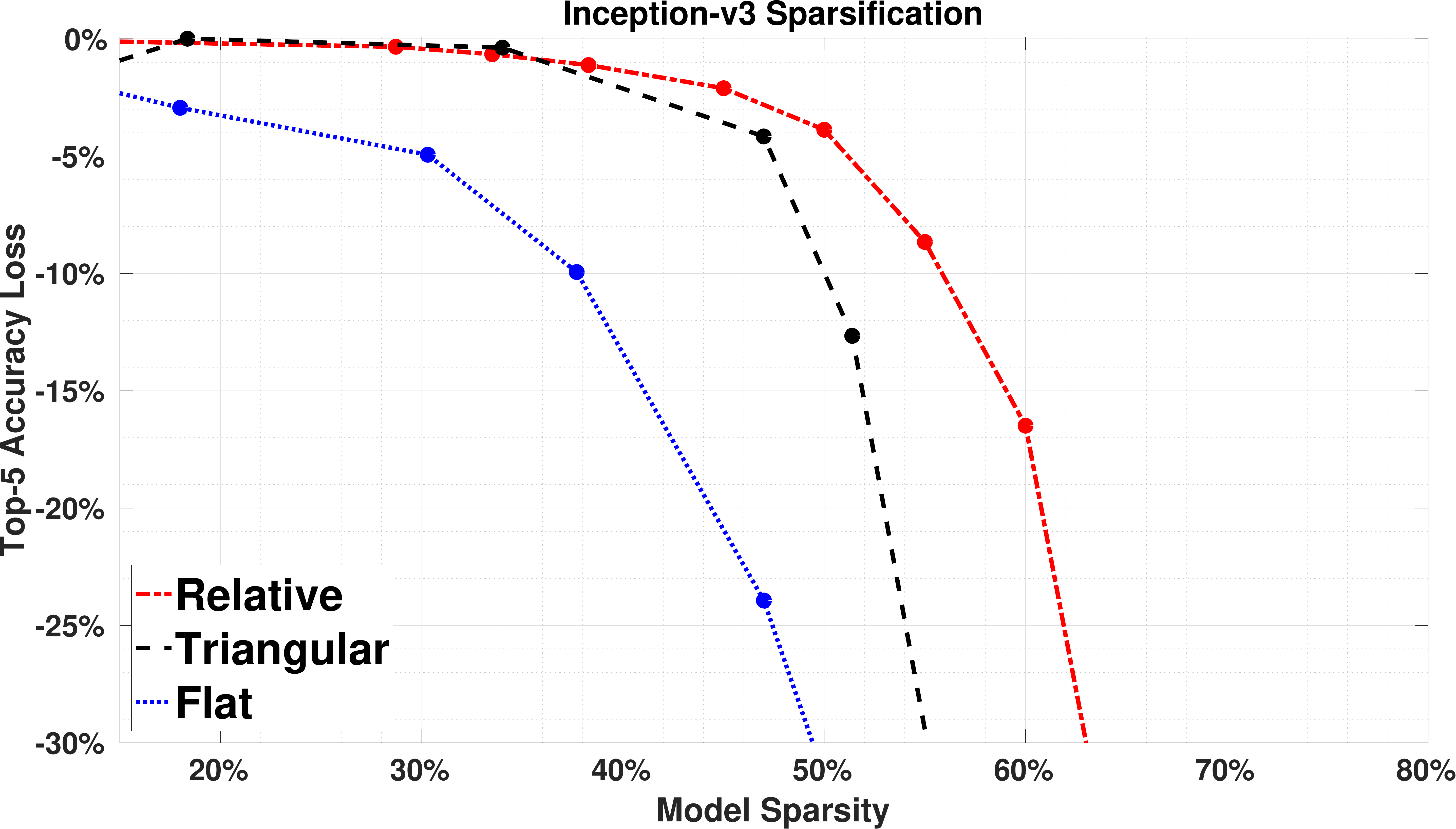}}}
\subfloat[MobileNets-v1]{
\label{fig:proposedSparsificationsL13_mobilenet}
\resizebox{0.46\textwidth}{!}{\includegraphics{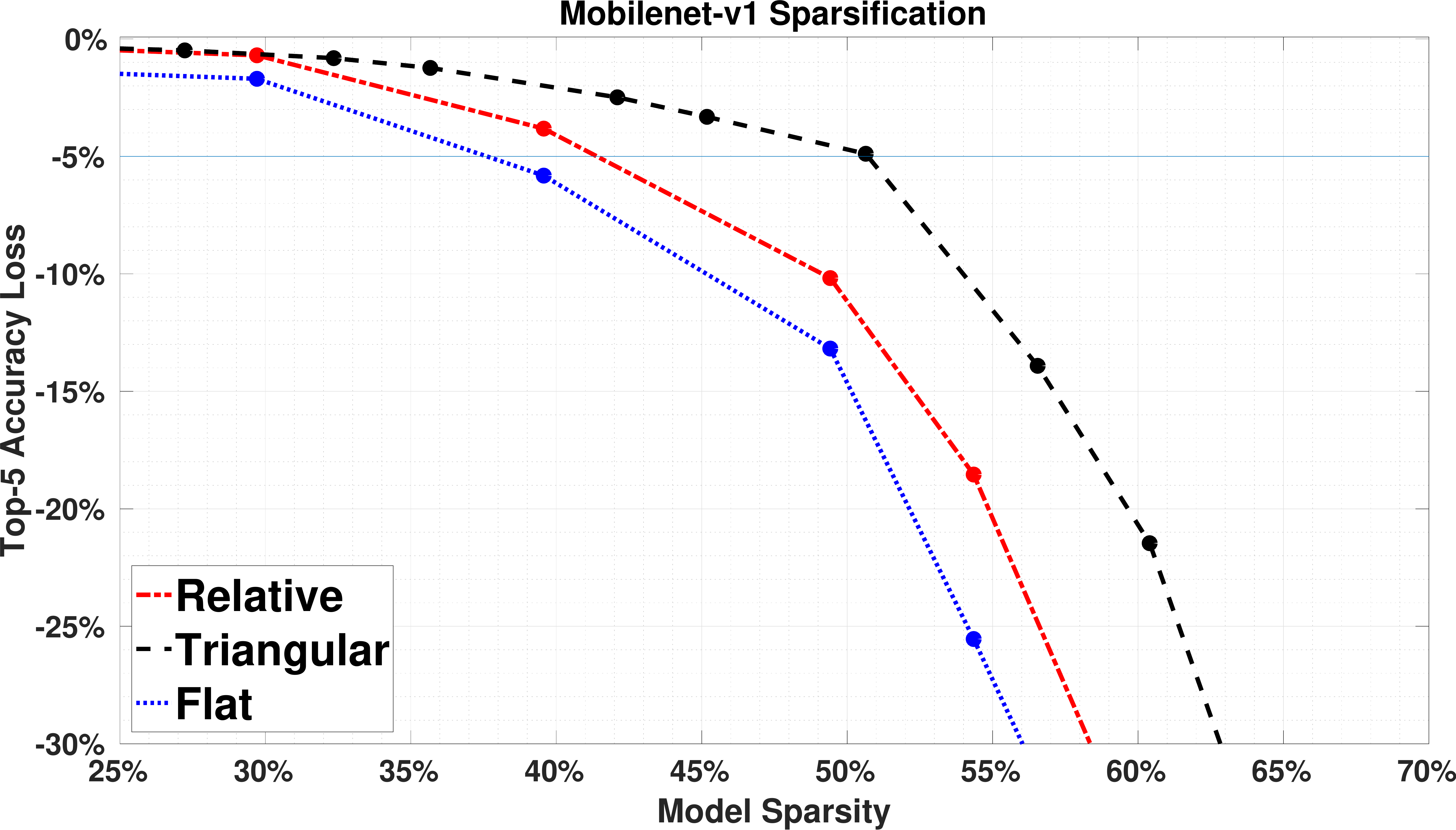}}}
\hfill
\subfloat[ResNet-v2]{
\label{fig:proposedSparsificationsL13_resnet}
\resizebox{0.46\textwidth}{!}{\includegraphics{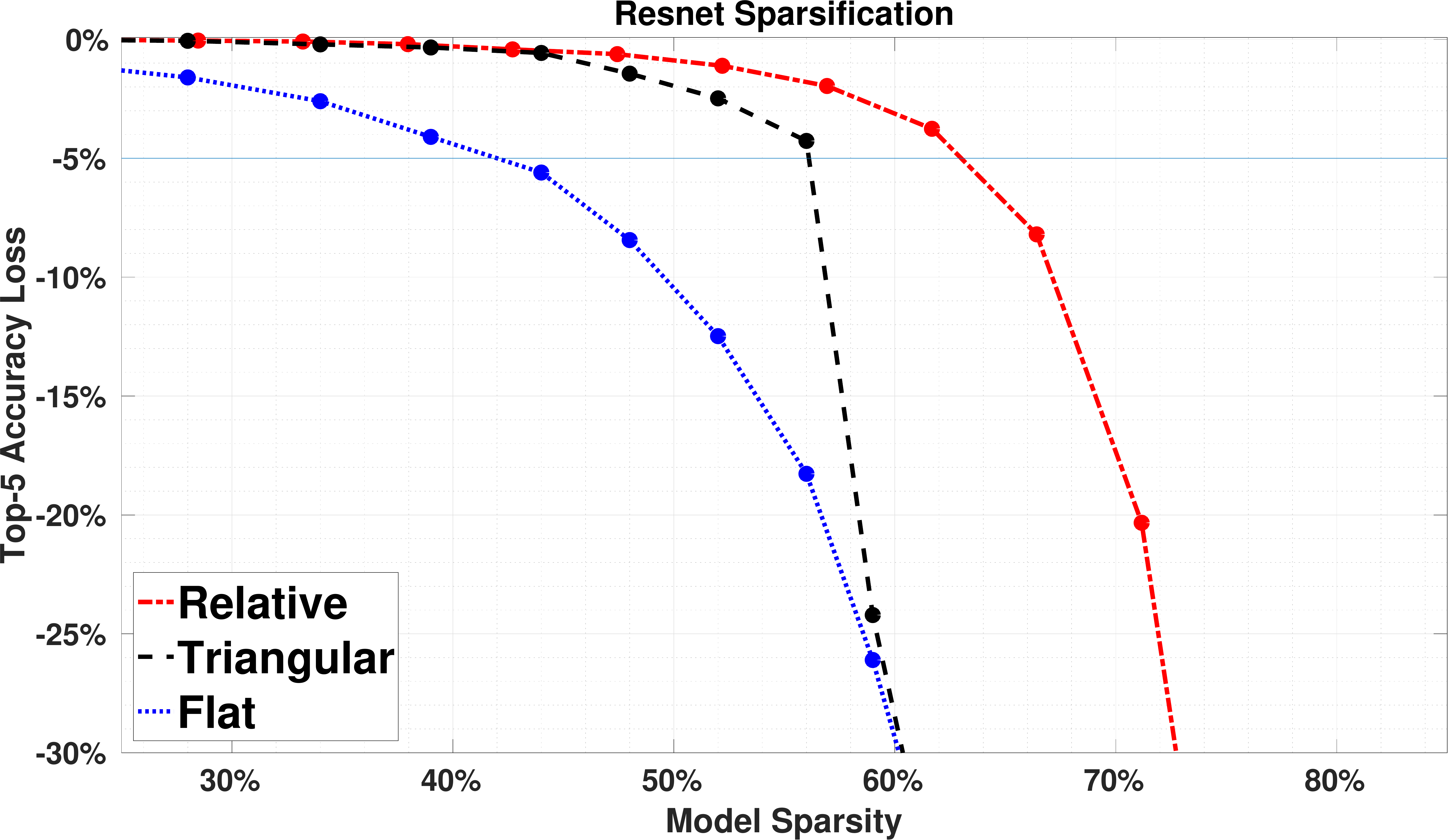}}}
\subfloat[VGG-16]{
\label{fig:proposedSparsificationsL13_vgg}
\resizebox{0.46\textwidth}{!}{\includegraphics{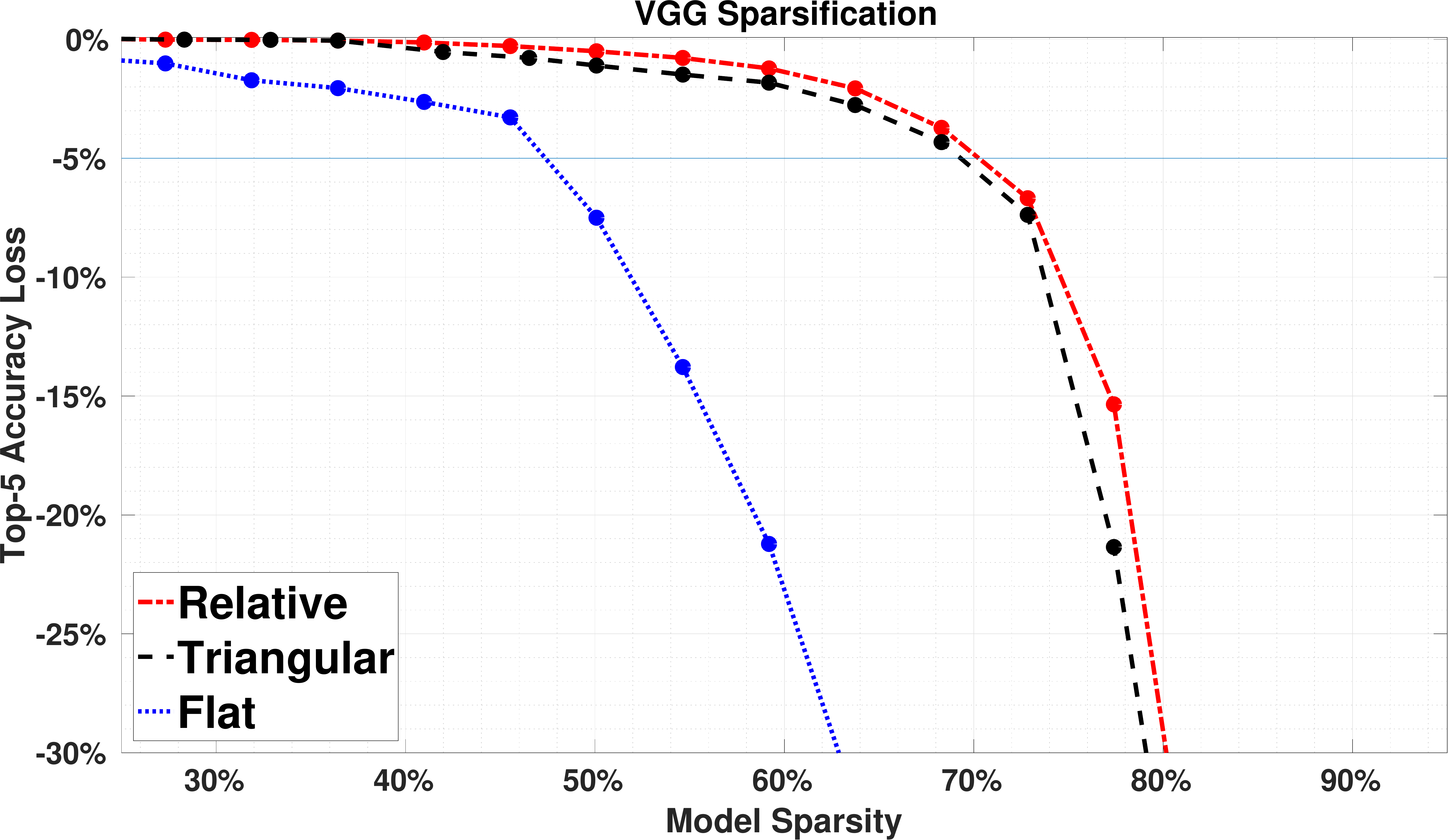}}}
\hfill
\hspace*{-1.3cm}
\subfloat[Alexnet]{
\label{fig:proposedSparsificationsL13_alexnet}
\resizebox{0.46\textwidth}{!}{\includegraphics{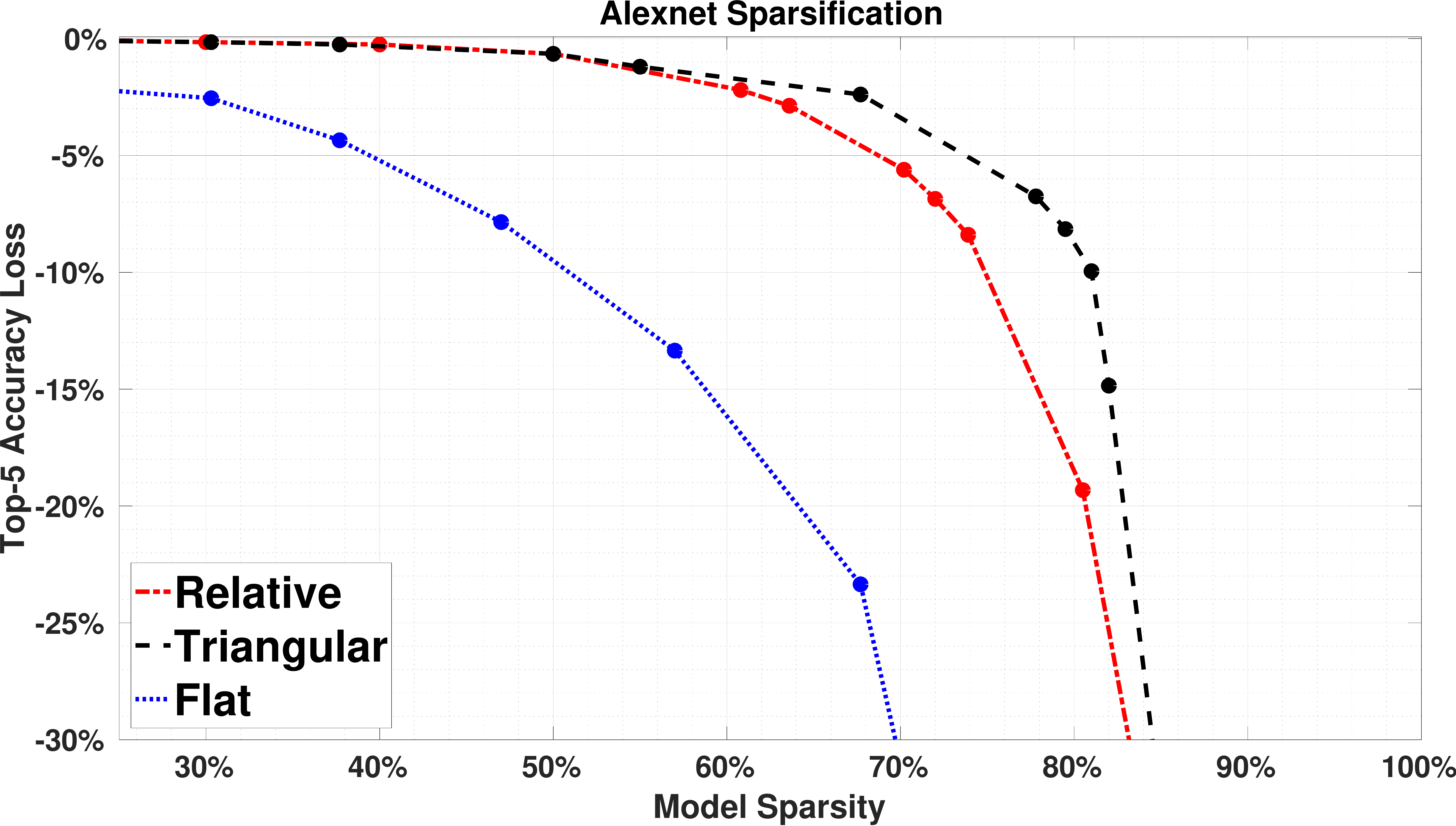}}} 
\hspace*{1cm} 
\subfloat[Alexnet Fine-tuning]{
\centering
    \label{tab:alexnetfinetuning}
    \resizebox{4cm}{!}{%
    \begin{tabular}[b]
    {| >{\centering\arraybackslash}m{1cm} | >{\centering\arraybackslash}m{1cm} | >{\centering\arraybackslash}m{1.5cm} |}
	\hline
	Layers & \#Params & \%Sparsified  \\ \hline
	Conv1  & 23 K & 10\% \\ \hline
	Conv2  & 307K & 35\% \\ \hline
	Conv3  & 663K & 35\% \\ \hline
	Conv4  & 1.3M &  35\% \\ \hline
	Conv5  & 884K & 35\% \\ \Xhline{2\arrayrulewidth}
	FC1  & 26M & 85\% \\ \hline
	FC2  & 16M & 85\% \\ \hline
	FC3  & 4M  & 73\% \\ \Xhline{2\arrayrulewidth}
	\textbf{Total}  & \textbf{50M}  & \textbf{81.1\%} \\ \hline
 	\end{tabular}}
 		}
\caption{Sparsity-Accuracy Trade-off for  Our Three Proposed Sparsification Methods} 
\end{figure}

Figures \ref{fig:proposedSparsificationsL30}, \ref{fig:proposedSparsificationsL13_mobilenet}, \ref{fig:proposedSparsificationsL13_resnet}, \ref{fig:proposedSparsificationsL13_vgg}, and  \ref{fig:proposedSparsificationsL13_alexnet} show the inference accuracy as a function of the introduced sparsity by each method. They reflect that significant sparsity can be obtained with a small reduction in inference accuracy. With less than 5\% reduction in accuracy, we gain 51\% sparsity (2.04$\times$ compression factor), 50\% (2$\times$), 62\% (2.63$\times$), 70\% (3.33$\times$), and 73\% (3.7$\times$) for the models. This validates our approach. 

Further, the figures reflect that the relative method outperforms the other two methods for the Inception-v3, VGG and ResNet, but the triangular one outperforms the other two for MobileNet-v1 
This is likely due to the structure of the models. MobileNet-v1 and Alexnet have a gradual linear increase in the size of the convolution layers, making the triangular method more effective. In contrast, the other models have no such increase, making the relative method more effective. As a case-in-point, ResNet has 152 \texttt{conv} layers with variable sizes, which  makes the triangular method less effective, as seen by the  drop in accuracy in Figure \ref{fig:proposedSparsificationsL13_resnet}. 

An interesting observation is that for Alexnet,  introducing the first 50\% sparsity suffers little drop in accuracy. This value is 35\%, 30\%, 41\%, and 42\% for the other models and it shows there exists significant redundancy within CNNs. Han et al. \cite{Han} observe the same with their Caffe implementation of Alexnet. The work was mainly focused on iteratively pruning and retraining CNNs to compensate the loss of accuracy. The authors' method with no retraining is not specified and it is unclear if it applies to other CNNs or not. However, the authors report a gain of around 80\% sparsity by pruning (i.e., without retraining) Alexnet with L2 regularization. Our evaluation validates their result across other models using the proposed on-the-fly methods. 

\textbf{Fine-tuning}. 
We explore what can be achieved by some fine-tuning of our methods, still with no retraining, in order to gain more sparsity. We do so to determine the effectiveness of our on-the-fly methods, since the fine-tuning is not likely feasible in out context. We focus on the relative method and start with a baseline sparsity. We then vary the degree of sparsity of each layer in turn around the base sparsity, attempting to maintain a no more than 5\% drop in inference accuracy.
The results for AlexNet are shown in the table in Figure~\ref{tab:alexnetfinetuning}. The baseline sparsity is selected as 70\%.
It is possible for some layers, particularly larger ones, to have higher sparsity, while smaller layers are more sensitive to sparsification and must have lower sparsity. Nonetheless, there is a gain of 8\% in overall model sparsity. This value is 4\%, 3\%, 2\% , and 5\% for Inception-v3, Mobilenet-v1, ResNet, and VGG, respectively. Since this gain comes at the expense of an exploration of different sparsity ratios for the layers and thus more computations, it is not feasible in the contexts we explore. However, the gain is not significant to render our on-the-fly methods inefficient on their own without further tuning.

\section{Concluding Remarks}
\label{sec:conc}

We explored sparsification without retraining of CNNs and proposed three model-independent methods doing so. We experimentally evaluated these methods and showed that they can result in up to 73\% sparsity with less than 5\% drop in inference accuracy. However, there is no single method that works best for all models.  Further, our evaluation showed that it is possible to fine-tune the methods to further gain sparsity with no significant drop in inference accuracy. However, such tuning of the methods is complex and suggests the need for autotuning. 
There are two key directions for future work. The first is to explore heuristics for selecting a sparsification method based on the CNN model and possibly fine tune the parameters of the methods.  Identifying the best sparsity configuration is important and the choice of the right sparsification method, the layers to target, and the sparsity ratio for each targeted layer presents an optimization problem. 
The second is to realize the benefit of the sparsity in the model's implementation on the NNlib library, which offloads neural networks operations from TensorFlow to Qualcomm's Hexagon-DSP. 
The introduction of zeros in the model weights can only benefit performance/energy if their presence is exploited. In particular, we also wish to address the additional challenge that arises in the quantized model, which stems from the fact that the representation of a zero is neither the value zero nor is unique. Thus, we will explore different quantization approaches to address this shortcoming.


\bibliography{bib}
\bibliographystyle{plain}

\end{document}